\def\year{2021}\relax
\documentclass[letterpaper]{article} 
\usepackage{aaai21}  
\usepackage{times}  
\usepackage{helvet} 
\usepackage{courier}  
\usepackage[hyphens]{url}  
\usepackage{graphicx} 
\usepackage{natbib}  
\usepackage{caption} 
\usepackage{bm}
\usepackage{amsmath}
\usepackage{subfigure}
\usepackage{amsfonts}
\usepackage{cuted}
\DeclareMathOperator*{\E}{\mathbb{E}}
\usepackage[switch]{lineno}

\title{Attention Forcing for Machine Translation}
\author{
    Qingyun Dou, Yiting Lu, Potsawee Manakul, Xixin Wu, Mark J. F. Gales \\
}
\affiliations{
    University of Cambridge\\
    \{qd212,ylt28,pm574,xw369,mjfg100\}@cam.ac.uk
}

\begin{document}

\maketitle

\begin{abstract}
Auto-regressive sequence-to-sequence models with attention mechanisms have achieved state-of-the-art performance in various tasks including Text-To-Speech (TTS) and Neural Machine Translation (NMT). The standard training approach, teacher forcing, guides a model with the reference output history. At inference stage, the generated output history must be used. This mismatch can impact performance. However, it is highly challenging to train the model using the generated output. Several approaches have been proposed to address this problem, normally by selectively using the generated output history. To make training stable, these approaches often require a heuristic schedule or an auxiliary classifier. This paper introduces attention forcing for NMT. This approach guides the model with the generated output history and reference attention, and can reduce the training-inference mismatch without a schedule or a classifier. Attention forcing has been successful in TTS, but its application to NMT is more challenging, due to the discrete and multi-modal nature of the output space. To tackle this problem, this paper adds a selection scheme to vanilla attention forcing, which automatically selects a suitable training approach for each pair of training data. Experiments show that attention forcing can improve the overall translation quality and the diversity of the translations.
\end{abstract}

\section{Introduction}
Auto-regressive sequence-to-sequence (seq2seq) models with attention mechanisms are used in a variety of areas including Neural Machine Translation (NMT) \cite{neubig2017neural,huang2016attention} and speech synthesis \cite{shen2018natural,wang2018style}, also known as Text-To-Speech (TTS). These models excel at connecting sequences of different length, but can be difficult to train. A standard approach is teacher forcing, which guides a model with reference output history during training. This makes the model unlikely to recover from its mistakes during inference, where the reference output is replaced by generated output. One alternative is to train the model in free running mode, where the model is guided by generated output history. This approach often struggles to converge, especially for attention-based models, which need to infer the correct output and align it with the input at the same time.

Several approaches have been introduced to tackle the above problem, namely scheduled sampling \cite{bengio2015scheduled} and professor forcing \cite{lamb2016professor}. Scheduled sampling randomly decides whether the reference or generated output token is added to the output history. The probability of choosing the reference output token decays with a heuristic schedule. Professor forcing views the seq2seq model as a generator. During training, the generator operates in both teacher forcing mode and free running mode. In teacher forcing mode, it tries to maximize the standard likelihood. In free running mode, it tries to fool a discriminator, which is trained to tell the mode of the generator.

This paper introduces attention forcing for NMT. This approach guides the model with the generated output history and reference attention, and can reduce the training-inference mismatch without a schedule or a classifier. Attention forcing has been successful in TTS, but its application to NMT is more challenging \cite{dou2019attention}. Analysis in this paper indicates that the difficulty comes from the discrete and multi-modal nature of the output space. To tackle this problem, a selection scheme is added to vanilla attention forcing: a suitable training approach is automatically selected for each pair of training data. Experiments show that attention forcing can improve the overall translation quality and the diversity of the translations, especially when the source and target languages are very different.

\section{Sequence-to-sequence Generation}
Sequence-to-sequence generation can be defined as mapping an input sequence $\bm{x}_{1:L}$ to an output sequence $\bm{y}_{1:T}$. From a probabilistic perspective, a model $\bm{\theta}$ estimates the distribution of $\bm{y}_{1:T}$ given $\bm{x}_{1:L}$, typically as a product of conditional distributions: $p(\bm{y}_{1:T}|\bm{x}_{1:L}; \bm{\theta}) = \textstyle \prod_{t=1}^{T} p(\bm{y}_{t}|\bm{y}_{<t},\bm{x}_{1:L}; \bm{\theta})$.

Ideally, the model is trained through minimizing the KL-divergence between the true distribution $p(\bm{y}_{1:T}|\bm{x}_{1:L})$ and the estimated distribution. In practice, this is approximated by minimizing the Negative Log-Likelihood (NLL) over some training data $\{\bm{y}^{*(n)}_{1:T}, \bm{x}^{(n)}_{1:L}\}_{1}^{N}$, sampled from the true distribution:
\begin{align}
\begin{split}
\mathcal{L}(\bm{\theta}) &= \E_{\scalebox{0.8}{$\bm{x}_{1:L}$}} \mathrm{KL} \big(p(\bm{y}_{1:T}|\bm{x}_{1:L}) || p(\bm{y}_{1:T}|\bm{x}_{1:L}; \bm{\theta}) \big) \\
&\propto - \textstyle \sum_{n=1}^{N} \log p(\bm{y}^{*(n)}_{1:T}|\bm{x}^{(n)}_{1:L}; \bm{\theta}) \label{eq:pb_train}
\end{split}
\end{align}
$\mathcal{L}(\bm{\theta})$ denotes the loss. To simplify the notation, the sum over data $\sum_{n=1}^{N}$ and the superscript index $^{(n)}$ will be omitted by default. During inference, given an input $\bm{x}_{1:L}$, the output can be obtained through searching for the most probable sequence from $p(\bm{y}_{1:T}| \bm{x}_{1:L}; \bm{\theta})$. The exact search is expensive and is often approximated by greedy search for continuous output, or beam search for discrete output \cite{bengio2015scheduled}.

\subsection{Attention-based Sequence-to-sequence Models}
\begin{figure}
\centering
\includegraphics[scale=0.28]{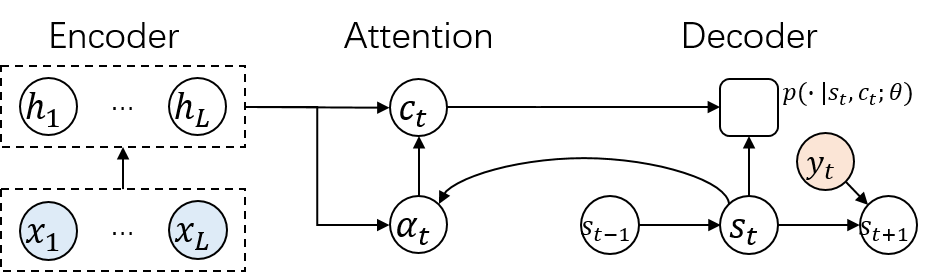}
\caption{Illustration of an attention-based encoder-decoder. The blue/orange nodes represent the input/output data. The monochrome nodes represent the variables generated from the model, among which the square node corresponds to the distribution of the output variable.}\label{fig:EAD}
\end{figure}

Attention mechanisms \cite{bahdanau2014neural,chorowski2015attention} are commonly used to connect sequences of different length. This paper focuses on attention-based encoder-decoder models. For these models, the probability $p(\bm{y}_{t}|\bm{y}_{<t}, \bm{x}_{1:L}; \bm{\theta})$ is estimated as:
\begin{align}
\begin{split}
p(\bm{y}_{t}|\bm{y}_{<t}, \bm{x}_{1:L}; \bm{\theta}) &\approx p(\bm{y}_{t}|\bm{y}_{<t}, \bm{\alpha}_{t}, \bm{x}_{1:L}; \bm{\theta}) \\
&\approx p(\bm{y}_{t}|\bm{s}_{t}, \bm{c}_{t}; \bm{\theta}_{y}) \label{eq:py_general_p} 
\end{split} 
\end{align}
$\bm{\theta} = \{\bm{\theta}_{y}, \bm{\theta}_{s}, \bm{\theta}_{c}\}$. $\bm{\alpha}_{t}$ is an alignment vector (a set of attention weights). $\bm{s}_{t}$ is a state vector representing the output history $\bm{y}_{<t}$, and $\bm{c}_{t}$ is a context vector summarizing $\bm{x}_{1:L}$ for the prediction of $\bm{y}_{t}$. Figure \ref{fig:EAD} and the following equations give a more detailed illustration of how $\bm{\alpha}_{t}$, $\bm{s}_{t}$ and $\bm{c}_{t}$ can be computed:
\begin{align}
\bm{h}_{1:L} &= f(\bm{x}_{1:L}; \bm{\theta}_{h}) \label{eq:EAD_enc}\\
\bm{s}_{t} &= f(\bm{s}_{t-1}, \bm{y}_{t-1}; \bm{\theta}_{s}) \label{eq:EAD_s}\\
\bm{\alpha}_{t} &= f(\bm{s}_{t}, \bm{h}_{1:L}; \bm{\theta}_{\alpha}) \label{eq:EAD_att_1}\\
\bm{c}_{t} &= \textstyle \sum_{l=1}^{L} \alpha_{t,l} \bm{h}_{l} \label{eq:EAD_att_2}\\
\bm{y}_{t} &\sim p(\cdot | \bm{s}_{t}, \bm{c}_{t}; \bm{\theta}_{y}) \label{eq:EAD_dec}
\end{align}
First the encoder maps $\bm{x}_{1:L}$ to an encoding sequence $\bm{h}_{1:L}$. For each decoder time step, $\bm{s}_{t}$ is updated with $\bm{y}_{t-1}$. Based on $\bm{h}_{1:L}$ and $\bm{s}_{t}$, the attention mechanism computes $\bm{\alpha}_{t}$, and then $\bm{c}_{t}$. Finally, the decoder estimates a distribution based on $\bm{s}_{t}$ and $\bm{c}_{t}$, and optionally generates an output token $\hat{\bm{y}_{t}}$. Note that while illustrated with this form of attention, attention forcing is not limited to it.


\subsection{Training Approaches} \label{sec:approaches}
As shown in equation \ref{eq:pb_train}, minimizing the KL-divergence can be approximated by minimizing the NLL. This motivates teacher forcing, where the reference output history is given to the model, and the loss can be written as:
\begin{equation} \label{eq:loss_y}
\begin{split}
    \mathcal{L}_{y}^{(\tt T)}(\bm{\theta}) &= - \textstyle  \sum_{t=1}^{T} \log p(\bm{y}^{*}_{t}| \bm{y}^{*}_{<t}, \bm{x}_{1:L}; \bm{\theta})
\end{split}
\end{equation}
This approach yields the correct model (zero KL-divergence) if the following assumptions hold: 1) the model is powerful enough ; 2) the model is optimized correctly; 3) there is enough training data to approximate the expectation shown in equation \ref{eq:pb_train}. However, these assumptions are often not true. In addition, at inference stage, the generated back history must be used. Hence the model is prone to mistakes that can accumulate across time.

In practice, the model is often assessed by some distance $\mathcal{D}$ between the reference $\bm{y}_{1:T}$ and the prediction $\hat{\bm{y}}'_{1:T}$. This motivates Minimun Bayes Risk (MBR) training, which minimizes the expectation of $\mathcal{D} (\bm{y}_{1:T}, \hat{\bm{y}}'_{1:T})$. This approach allows directly optimizing $\mathcal{D}$ \cite{ranzato2015sequence, bahdanau2016actor}. Note that Reinforcement Learning (RL) is a type of MBR training. $\mathcal{D}$ does not need to be differentiable, and $\bm{y}_{1:T}$ and $\hat{\bm{y}}'_{1:T}$ do not need to be aligned. However, for many tasks such as TTS and NMT, there is no gold-standard distance metric other than subjective human evaluation.

Although defined for sequences, $\mathcal{D}$ is usually computed at sub-sequence level, e.g. BLEU score for NMT and $L_{p}$ distance for TTS. So training the model to predict the reference output, based on erroneous output history, indirectly reduces the Bayes risk. One example is to train the model in free running mode, where the generated output history is given to the model, and the probability term in equation \ref{eq:loss_y} becomes $p(\bm{y}^{*}_{t}| \hat{\bm{y}}'_{<t}, \bm{x}_{1:L}; \bm{\theta})$. This approach often struggles to converge, and several approaches are proposed to tackle this problem, namely scheduled sampling and professor forcing.

Scheduled sampling \cite{bengio2015scheduled} randomly decides whether the reference or generated output is added to the history. For this approach, the probability term in equation \ref{eq:loss_y} becomes $p(\bm{y}^{*}_{t}| \widetilde{\bm{y}}_{<t}, \bm{x}_{1:L}; \bm{\theta})$; $\widetilde{\bm{y}}_{t}=\bm{y}_{t}$ with probability $\epsilon$, and $\hat{\bm{y}}'_{t}$ otherwise. $\epsilon$ gradually decays from 1 to 0 with a heuristic schedule. This approach improves the results in many cases \cite{bengio2015scheduled}, but sometimes lead to worse results \cite{bengio2015scheduled, wang2017tacotron, guo2019new}. One concern is the decay schedule not fitting the learning pace of the model, another is that $\widetilde{\bm{y}}_{<t}$ is usually an inconsistent mixture of the reference and generated output.
Professor forcing \cite{lamb2016professor} is an alternative approach. During training, the model generates two sequences for each input sequence, respectively in teacher forcing mode and free running mode\footnote{The term "teacher forcing", as well as "attention forcing", can refer to either an operation mode, or the approach to train a model in that operation mode. An operation mode can be used not only to train a model, but also to generate from it.}. The output and/or some hidden sequences are used to train a discriminator, which estimates the probability that a group of sequences is generated in teacher forcing mode. For the generator, there are two training objectives: 1) the standard NLL loss; 2) to fool the discriminator in free running mode, and optionally in teacher forcing mode. This approach regularizes the output and/or some hidden layers, encouraging them to behave as if in teacher forcing mode, at the expense of tuning the discriminator.

To our knowledge, teacher forcing is the most standard training approach for many seq2seq tasks such as TTS and NMT. Despite its drawbacks, using the reference output history usually makes it easy for training to converge, and in some cases, the drawbacks can be alleviated without changing the training approach. For example, one side effect of teacher forcing is that there might be too much information in the output history, to the extent that the model does not take enough information from the input when predicting the next token. For TTS, the model might learn to copy the previous token, as adjacent tokens are often similar; for NMT, the model might behave like a language model, overlooking the input. A natural solution is to reduce the frame rate, i.e. to downsamlpe the output sequences. While straight-forward for TTS, this is difficult to apply to NMT, due to the discrete nature of the output space.




\section{Attention Forcing}
\subsection{Vanilla Attention Forcing} \label{sec:vaf}
The basic idea of attention forcing is to use the reference attention and generated output to guide the model during training. In attention forcing mode, the model does not need to learn to simultaneously infer the output and align it with the input. As the reference alignment is known, the decoder can focus on inferring the output, and the attention mechanism can focus on generating the correct alignment.

\begin{figure}
\centering
\includegraphics[scale=0.28]{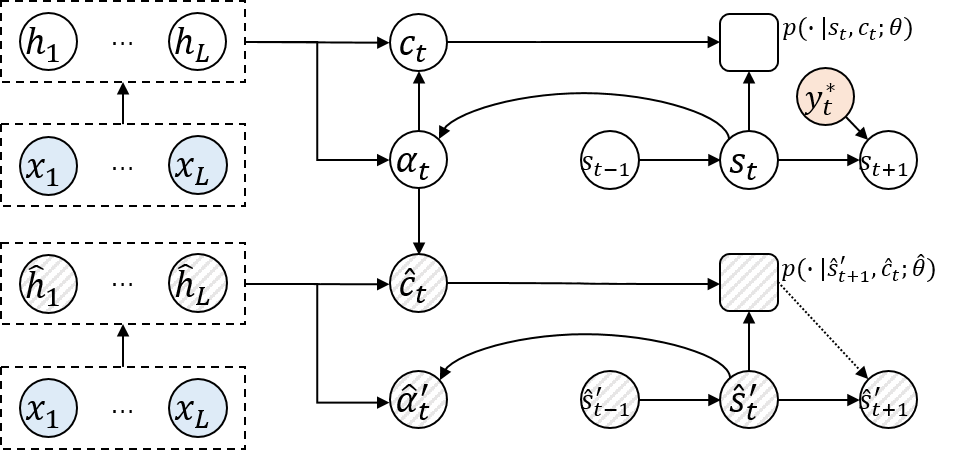}
\caption{Illustration of vanilla attention forcing. The white nodes represent the variables from the teacher forcing model, and the shaded nodes the attention forcing model.}
\label{fig:AF}
\end{figure}

Let $\hat{\bm{\theta}}$ denote the model trained with attention forcing, and later used for inference. Let $\hat{\bm{y}}_{1:T}$ / $\hat{\bm{y}}'_{1:T}$ denote $\hat{\bm{\theta}}$'s output, guided by the reference / generated output. The convention in this paper is that if the generated output is used to produce a variable, a prime ($'$) will be added to its notation. In attention forcing mode, the probability $p(\bm{y}^{*}_{t}|\bm{y}^{*}_{<t}, \bm{x}_{1:L}; \hat{\bm{\theta}})$ is estimated with the generated output $\hat{\bm{y}}'_{<t}$ and the reference alignment $\bm{\alpha}_{t}$:
\begin{equation} \label{eq:py_A}
\begin{split}
p(\bm{y}^{*}_{t}|\bm{y}^{*}_{<t}, \bm{x}_{1:L}; \hat{\bm{\theta}}) &\approx p(\bm{y}^{*}_{t}| \hat{\bm{y}}'_{<t}, \bm{\alpha}_{t}, \bm{x}_{1:L}; \hat{\bm{\theta}}) \\
&\approx p(\bm{y}^{*}_{t} | \hat{\bm{s}}'_{t}, \hat{\bm{c}}_{t}; \hat{\bm{\theta}}_{y})
\end{split}
\end{equation}
$\hat{\bm{s}}'_{t}$ and $\hat{\bm{c}}_{t}$ denote the state vector and context vector generated by $\hat{\bm{\theta}}$. The context vector does not depend on the generated output, hence the notation $\hat{\bm{c}}_{t}$ instead of $\hat{\bm{c}}'_{t}$. Details of vanilla attention forcing are illustrated by figure \ref{fig:AF}, as well as the following equations:
\begin{align}
&\bm{h}_{1:L} = f(\bm{x}_{1:L}; \bm{\theta}_{h}) & &\hat{\bm{h}}_{1:L} = f(\bm{x}_{1:L}; \hat{\bm{\theta}}_{h}) \label{eq:AF_enc}\\
&\bm{s}_{t} = f(\bm{s}_{t-1}, \bm{y}^{*}_{t-1}; \bm{\theta}_{s}) & &\hat{\bm{s}}'_{t} = f(\hat{\bm{s}}'_{t-1}, \hat{\bm{y}}'_{t-1}; \hat{\bm{\theta}}_{s}) \label{eq:AF_s}\\
&\bm{\alpha}_{t} = f(\bm{s}_{t}, \bm{h}_{1:L}; \bm{\theta}_{\alpha}) & &\hat{\bm{\alpha}}'_{t} = f(\hat{\bm{s}}'_{t}, \hat{\bm{h}}_{1:L}; \hat{\bm{\theta}}_{\alpha}) \label{eq:AF_att_1}
\end{align}
\begin{align}
\hat{\bm{c}}_{t} &=\textstyle \sum_{l=1}^{L} \alpha_{t,l} \hat{\bm{h}}_{l} \label{eq:AF_att_2}\\ 
\hat{\bm{y}}'_{t} &\sim p(\cdot | \hat{\bm{s}}'_{t}, \hat{\bm{c}}_{t}; \hat{\bm{\theta}}_{y}) \label{eq:AF_dec} 
\end{align}

The right side of equations \ref{eq:AF_enc} to \ref{eq:AF_att_1}, as well as equations \ref{eq:AF_att_2} and \ref{eq:AF_dec}, show how the attention forcing model $\hat{\bm{\theta}}$ operates. $\hat{\bm{s}}'_{t}$ is computed with $\hat{\bm{y}}'_{<t}$. While an alignment $\hat{\bm{\alpha}}'_{t}$ is generated by $\hat{\bm{\theta}}$, it is not used by the decoder, because $\hat{\bm{c}}_{t}$ is computed with the reference alignment $\bm{\alpha}_{t}$. In most cases, $\bm{\alpha}_{t}$ is not available. One option of obtaining it is shown by the left side of equations \ref{eq:AF_enc} to \ref{eq:AF_att_1}: to generate $\bm{\alpha}_{t}$ from a teacher forcing model $\bm{\theta}$. $\bm{\theta}$ is trained in teacher forcing mode, and generates $\bm{\alpha}_{t}$ in the same mode.

During training, there are two objectives: to infer the reference output and to imitate the reference alignment. The respective loss functions are:
\begin{align}
\mathcal{L}_{y}^{(\tt A)}(\hat{\bm{\theta}}) &= -\scalebox{1.0}{$ \sum_{t=1}^{T}$} \log p(\bm{y}{^{*}_{t}}|\hat{\bm{y}}{'_{<t}}, \bm{\alpha}{_{t}}, \bm{x}{_{1:L}}; \hat{\bm{\theta}}) \label{eq:loss_y_AF}\\ 
\mathcal{L}_{\alpha}^{(\tt A)} (\hat{\bm{\theta}}) &= \scalebox{1.0}{$ \sum_{t=1}^{T}$} \mathrm{KL}(\bm{\alpha}_{t}||\hat{\bm{\alpha}}_{t}') \label{eq:loss_align}
\end{align}
As an alignment corresponds to a categorical distribution, KL-divergence is a natural difference metric. The two losses can be jointly optimized as $\mathcal{L}_{y,\alpha}^{(\tt A)} = \mathcal{L}_{y}^{(\tt A)} + \gamma \mathcal{L}_{\alpha}^{(\tt A)}$. $\gamma$ is a scaling factor that should be set according to the dynamic range of the two losses. Our default optimization option is as follows. $\bm{\theta}$ is trained in teacher forcing mode, and then fixed to generate the reference attention. $\hat{\bm{\theta}}$ is trained with the joint loss $\mathcal{L}_{y,\alpha}^{(\tt A)}$. This option makes training more stable, most probably because the reference attention is the same in each epoch. An alternative is to train $\bm{\theta}$ and $\hat{\bm{\theta}}$ simultaneously to save time. Another is to tie (parts of) $\bm{\theta}$ and $\hat{\bm{\theta}}$ to save memory.

At inference stage, the attention forcing model operates in free running mode. In this case, equations \ref{eq:AF_att_2} and \ref{eq:AF_dec} become $\hat{\bm{c}}'_{t} = \textstyle \sum_{l=1}^{L} \hat{\alpha}'_{t,l} \hat{\bm{h}}_{l}$ and $\hat{\bm{y}}'_{t} \sim p(\cdot | \hat{\bm{s}}'_{t}, \hat{\bm{c}}'_{t}; \hat{\bm{\theta}}_{y})$. The decoder is guided by $\hat{\bm{\alpha}}'_{t}$, instead of $\bm{\alpha}_{t}$.

\subsection{Challenges of the NMT Output Space} \label{sec:issues}

\begin{figure*}%
    \centering
    \subfigure{{\includegraphics[height=1.8cm]{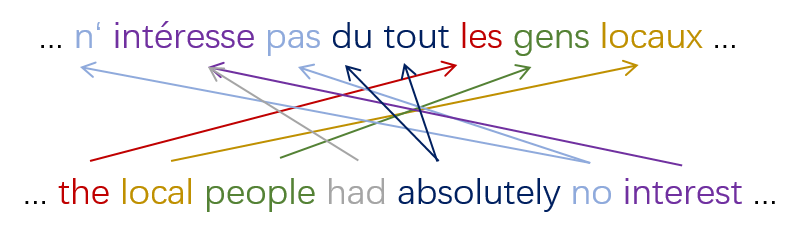} }}%
    \qquad
    \subfigure{{\includegraphics[height=1.8cm]{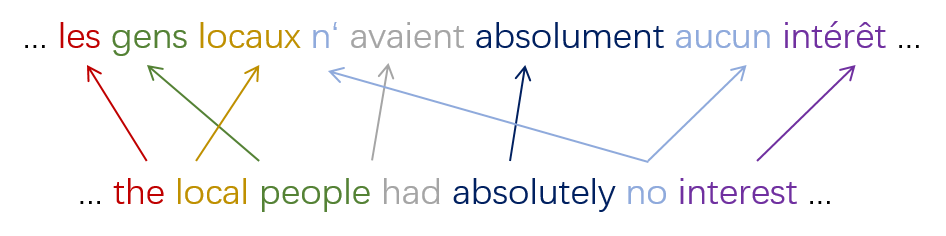} }}%
    \caption{Manual alignments between: left) the input and the reference output; right) the input and the generated output.} \label{align-TF-AF-manual}
\end{figure*}
\begin{figure*}%
    \centering
    \subfigure{{\includegraphics[width=5.5cm]{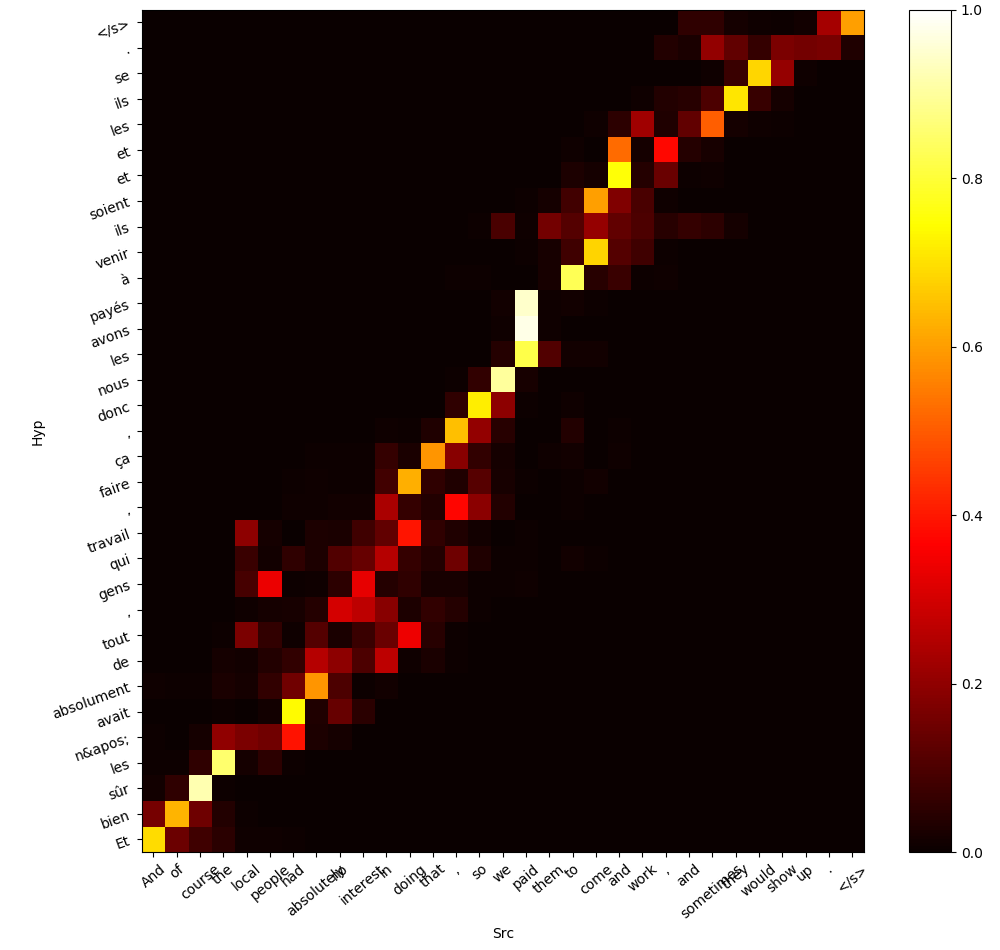} }}%
    \qquad \qquad 
    \subfigure{{\includegraphics[width=5.5cm]{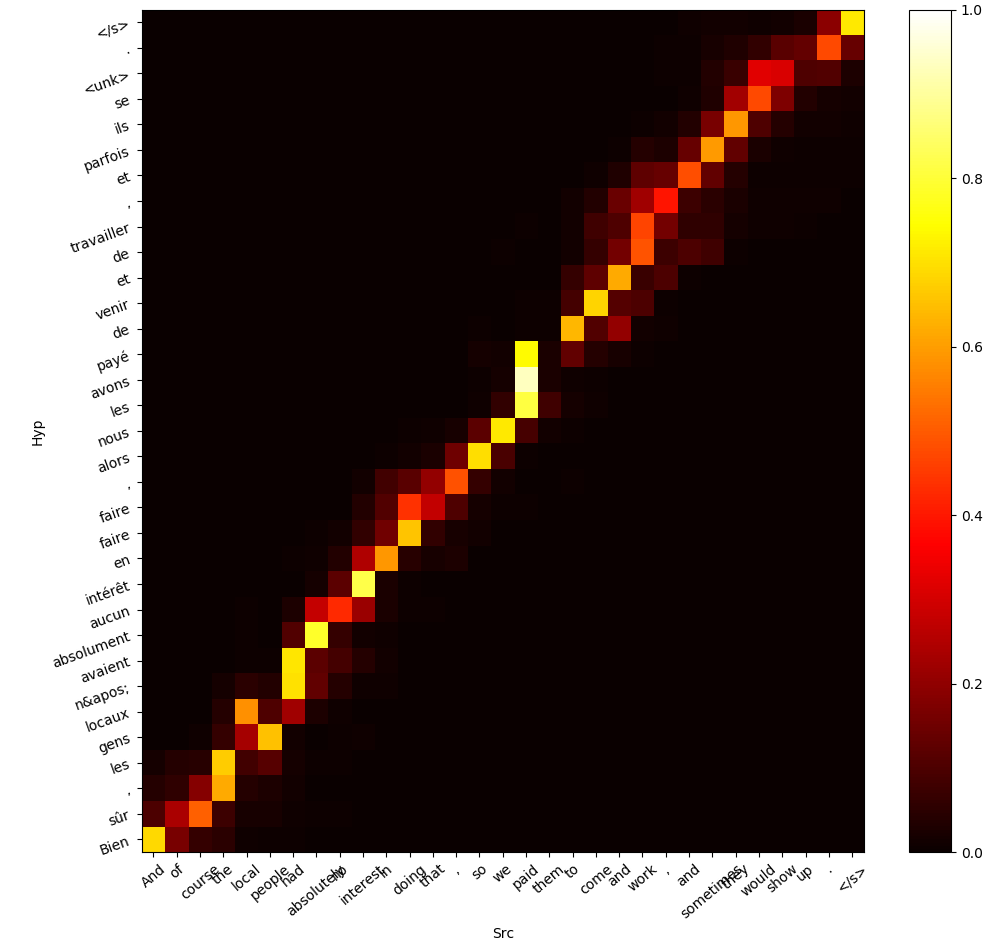} }}%
    \caption{Model-generated alignments using: left) the reference back history; right) the generated back history.} \label{align-TF-AF}
\end{figure*}

Previous research has shown that the vanilla form of attention forcing yields significant performance gain for TTS, but not for NMT \cite{dou2019attention}. Although both TTS and NMT are seq2seq tasks, they have considerable differences. For TTS, the attention is monotonic, which is relatively easy to learn. For NMT, the attention is much more complicated: there may be multiple valid modes of attention. The ordering of tokens can be changed while the meaning of the sequence remains the same. If the model takes an ordering that is different from the reference output, the token-level losses will mislead both the output and the attention. To illustrate the problem, consider the following example from our preliminary experiments on English-to-French NMT. For the input "... the local people had absolutely no interest ...", the reference output is "... n'intéresse pas du tout les gens locaux ...". When using the generated output history, the model outputs "... les gens locaux n'avaient absolument aucun intérêt ...". Figure \ref{align-TF-AF-manual} shows the manual alignments in these two cases. Figure \ref{align-TF-AF} shows the model-generated alignments, respectively using the reference and generated back history.\footnote{The complete input is "And of course the local people had absolutely no interest in doing that, so we paid them to come and work, and sometimes they would show up."} Obviously the alignment corresponding to the reference output is not a sensible target for the generated output.

Furthermore, even if the model happens to follow the reference ordering, there might be other issues. The output of TTS is usually orders of magnitude longer than that of NMT. For each time step, TTS has a continuous one-dimensional output space, while NMT has a discrete high-dimensional output space. Due to the length and sparsity of the NMT output, the errors in the output history tend to be more serious, to the extent that the token-level target is not appropriate. To illustrate the problem, suppose the reference output is "Thank you for listening", and the model predicts "Thanks" at the first time step. In this case, the next output should not be "you". "For" would instead be a more sensible target.


\subsection{Automatic Attention Forcing}

To tackle the above issues, we introduce automatic attention forcing. The basic idea is to automatically decide, for each pair of training data, whether vanilla attention forcing will be used. This is realized by tracking the alignment between the reference and the predicted outputs. If they are relatively well-aligned, vanilla attention forcing will be used, otherwise a more stable training mode will be used.

\begin{figure}
\centering
\includegraphics[scale=0.28]{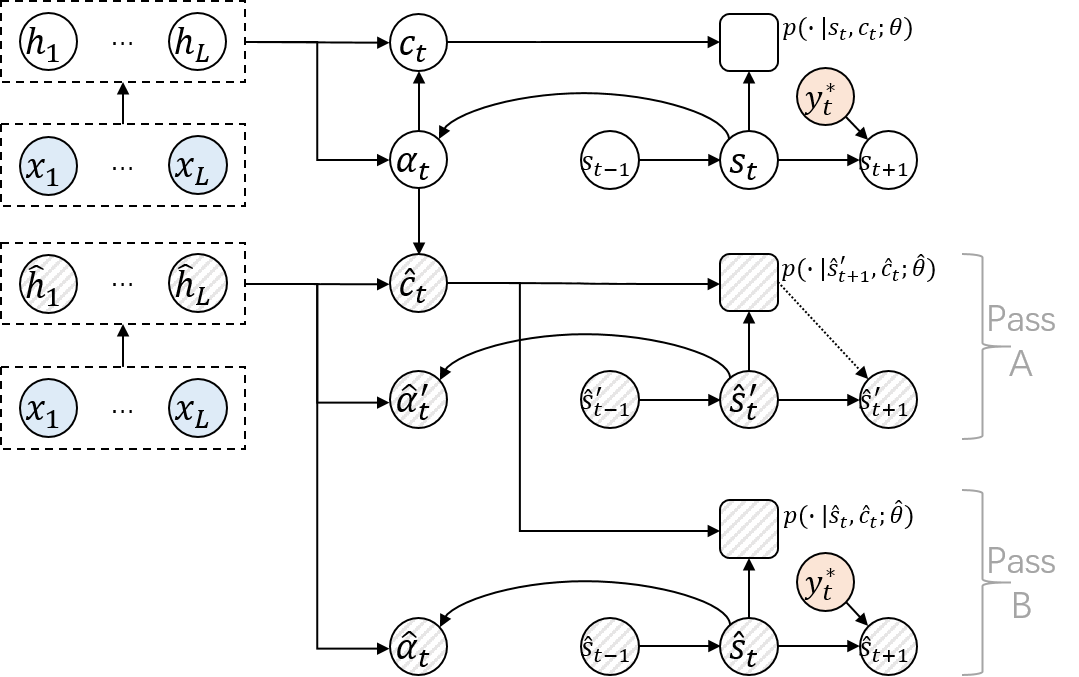}
\caption{Illustration of automatic attention forcing. Passes A and B share the same model parameters.} \label{fig:AAF}
\end{figure}

For each input sequence, the attention forcing model $\hat{\bm{\theta}}$ takes two forward passes. As illustrated by figure \ref{fig:AAF}, pass A is guided by the generated output history $\hat{\bm{y}}'_{<t}$, and pass B the reference output history $\bm{y}^{*}_{<t}$. The two forward passes can be completed in parallel, so no extra computation time is necessary. This can be formulated as follows.

\begin{align}
&\bm{h}_{1:L} = f(\bm{x}_{1:L}; \bm{\theta}_{h}) & &\hat{\bm{h}}_{1:L} = f(\bm{x}_{1:L}; \hat{\bm{\theta}}_{h}) \label{eq:AAF_enc}\\
&\bm{s}_{t} = f(\bm{s}_{t-1}, \bm{y}^{*}_{t-1}; \bm{\theta}_{s}) & &\begin{matrix}
\hat{\bm{s}}'_{t} = f(\hat{\bm{s}}'_{t-1}, \hat{\bm{y}}'_{t-1}; \hat{\bm{\theta}}_{s})\\ 
\hat{\bm{s}}_{t} = f(\hat{\bm{s}}_{t-1}, \bm{y}^{*}_{t-1}; \hat{\bm{\theta}}_{s})
\end{matrix} \\
&\bm{\alpha}_{t} = f(\bm{s}_{t}, \bm{h}_{1:L}; \bm{\theta}_{\alpha}) & &\begin{matrix}
\hat{\bm{\alpha}}'_{t} = f(\hat{\bm{s}}'_{t}, \hat{\bm{h}}_{1:L}; \hat{\bm{\theta}}_{\alpha})\\ 
\hat{\bm{\alpha}}_{t} = f(\hat{\bm{s}}_{t}, \hat{\bm{h}}_{1:L}; \hat{\bm{\theta}}_{\alpha})
\end{matrix}
\end{align}
\begin{align}
&\hat{\bm{c}}_{t} =\textstyle \sum_{l=1}^{L} \alpha_{t,l} \hat{\bm{h}}_{l} \\
&\begin{matrix}
\hat{\bm{y}}'_{t} \sim p(\cdot | \hat{\bm{s}}'_{t}, \hat{\bm{c}}_{t}; \hat{\bm{\theta}}_{y})\\ 
\hat{\bm{y}}_{t} \sim p(\cdot | \hat{\bm{s}}_{t}, \hat{\bm{c}}_{t}; \hat{\bm{\theta}}_{y})
\end{matrix}
\end{align}

Next, the choice of training mode is made at the sequence level, which ensures the consistency of the output history. If $\textstyle k \sum_{t=1}^{T} \mathrm{KL}( \bm{\alpha}_{t} || \hat{\bm{\alpha}}_{t}; \hat{\bm{\theta}}) > \sum_{t=1}^{T} \mathrm{KL}( \bm{\alpha}_{t} || \hat{\bm{\alpha}}'_{t}; \hat{\bm{\theta}})$, meaning that $\hat{\bm{y}}'_{1:T}$ is well aligned with $\bm{y}^{*}_{1:T}$, forward pass A will be used in the backpropagation. The loss is the same as in vanilla attention forcing:
\begin{align}
\begin{split}
\mathcal{L}_{y,\alpha}^{(\tt A)}(\hat{\bm{\theta}}) &= \textstyle \sum_{t=1}^{T} \log p(\bm{y}^{*}_{t} | \bm{x}_{1:L}, \hat{\bm{y}}'_{<t}, \bm{\alpha}_{t}; \hat{\bm{\theta}}) \\
&+ \textstyle \gamma \sum_{t=1}^{T} \mathrm{KL}( \bm{\alpha}_{t} || \hat{\bm{\alpha}}'_{t}; \hat{\bm{\theta}}) )
\end{split}
\end{align}

Otherwise forward pass B will be used:
\begin{align}
\begin{split}
\mathcal{L}_{y,\alpha}^{(\tt A)}(\hat{\bm{\theta}}) &= \textstyle \sum_{t=1}^{T} \log p(\bm{y}^{*}_{t} | \bm{x}_{1:L}, \bm{y}^{*}_{<t}, \bm{\alpha}_{t}; \hat{\bm{\theta}}) \\
&+ \textstyle \gamma \sum_{t=1}^{T} \mathrm{KL}( \bm{\alpha}_{t} || \hat{\bm{\alpha}}_{t}; \hat{\bm{\theta}}) )
\end{split}
\end{align}

The KL attention loss is used to determine if the alignment is good enough between the reference output $\bm{y}^{*}_{1:T}$ and the free running output $\hat{\bm{y}}'_{1:T}$. As both $\bm{\alpha}_{t}$ and $\hat{\bm{\alpha}}_{t}$ are computed using $\bm{y}^{*}_{<t}$, they are expected to be similar, yielding a relatively small $\mathrm{KL}( \bm{\alpha}_{t} || \hat{\bm{\alpha}}_{t}; \hat{\bm{\theta}})$. In contrast, $\hat{\bm{\alpha}}'_{t}$ is computed using $\hat{\bm{y}}'_{<t}$, and $\mathrm{KL}( \bm{\alpha}_{t} || \hat{\bm{\alpha}}'_{t}; \hat{\bm{\theta}})$ is expected to be larger. The hyper parameter $k$ is introduced to determine how out-of-alignment $\hat{\bm{y}}'_{1:T}$ and $\bm{y}^{*}_{1:T}$ can be. If $k=+\infty$, automatic attention forcing will be the same as vanilla attention forcing. In the following section, unless otherwise stated, the term "attention forcing" will refer to automatic attention forcing.

An empirical issue is that as training progresses, most elements in the alignment vectors will be close to 0. So the KL attention loss, which involves the log of the elements, might be unstable. To improve numerical stability, a trick is to smooth all alignment vectors with a uniform distribution $\bm{u}$. For example, $\mathrm{KL}( \bm{\alpha}_{t} || \hat{\bm{\alpha}}_{t}; \hat{\bm{\theta}})$ can be smoothed as $\mathrm{KL}( (1-\epsilon) \bm{\alpha}_{t} + \epsilon \bm{u} || (1-\epsilon) \hat{\bm{\alpha}}_{t} + \epsilon \bm{u}; \hat{\bm{\theta}} )$. $\epsilon$ is a small constant balancing the stability and accuracy of training; it is set to $e^{-10}$ in our experiments.

\subsection{Comparison with Related Work}
Intuitively, attention forcing, as well as scheduled sampling and professor forcing, sits between teacher forcing and free running. An advantage of attention forcing is that it does not require a schedule or a discriminator, which can be difficult to tune. Variations of scheduled sampling have been applied to NMT \cite{zhang2019bridging, duckworth2019parallel}; while \cite{zhang2019bridging} finds it helpful, \cite{duckworth2019parallel} reports slightly worse performance. Similarly, in TTS, both positive \cite{liu2020teacher} and negative \cite{guo2019new} results have been reported, showing that the schedule can be hard to tune.

In terms of regularization, attention forcing is similar to professor forcing. The output layer of the attention mechanism is regularized, and the KL-divergence is a well-established difference metric. \cite{guo2019new} and \cite{liu2020teacher} also perform hidden layer regularization, and both regularize the decoder states, for which there is not a natural difference metric. \cite{guo2019new} introduces a specific discriminator; \cite{liu2020teacher} experiments with $L_{1}$ loss, and one concern is the implicit assumption that the states are in $L_{1}$ space. 
The effect of regularization on attention mechanisms has been studied in previous work \cite{yu2017supervising,liu-etal-2016-neural,bao2018deriving}, where alternative approaches of obtaining reference attention are introduced. \cite{bao2018deriving} and \cite{yu2017supervising} require collecting extra data for reference attention, and \cite{liu-etal-2016-neural} uses a statistical machine translation model to estimate them. In contrast, we propose to generate the reference attention with a teacher forcing model, which can be trained simultaneously with the attention forcing model.

While this paper focuses on the discrete and multi-modal output space of NMT, attention forcing can be applied to both continuous and discrete outputs. Beam Search Optimization (BSO) \citep{wiseman2016sequence} is a NMT-specific alternative. To deal with the training-inference mismatch, it approximates beam search during training and penalizes the reference output token falling off the beam. This limits the output space to be discrete, so that beam search can be used. RL, a type of MBR, has also been applied to NMT \cite{ranzato2015sequence, bahdanau2016actor}. It trains the model in free running mode, and allows using a sequence-level loss. As discussed previously, a problem is that NMT does not have a gold-standard objective metric. The model trained with one metric (e.g. BLEU) might not excel at another (e.g. ROUGE) \cite{ranzato2015sequence}. To tackle this issue, adversarial training has been combined with RL \cite{yu2017seqgan, wu2018adversarial}: a discriminator learns a loss function, which is potentially better than standard metrics. For NMT, an additional motivation is that the discrete output is not differentiable, so RL is essential for training. Similar to the case of professor forcing, the discriminator itself can be difficult to train \cite{zhang2018bidirectional}.


A challenge of attention forcing is that when applying it to models without an attention mechanism, attention needs to be defined first. For convolutional neural networks, for example, attention maps can be defined based on the activation or gradient \cite{zagoruyko2016paying}. Some recent work on TTS \cite{ren2019fastspeech, ren2020fastspeech, yu2019durian, donahue2020end} uses a duration model instead of attention. In this case, one-hot alignment vectors can be defined according to the duration of input tokens.

\section{Experiments}
\subsection{Experimental Setup}
\subsubsection{Data}
The experiments are conducted on two NMT tasks: English-to-French (EnFr) and English-to-Vietnamese (EnVi) in IWSLT 2015. For EnFr, the training set contains 208k sentence pairs. The validation set is tst2013 and the test set tst2014. For EnVi, the training set contains 133k sentence pairs. The validation set is tst2012 and the test set tst2013. The latter is expected to be a harder task, considering the differences between the source and target languages in terms of syntax and lexicon.

\subsubsection{Model, Training and Inference}
The model is similar to Google's attention-based encoder-decoder LSTM \cite{wu2016google}. The differences are as follows. The model is simplified with a smaller number of LSTM layers due to the small scale of data: the encoder has 2 layers of bi-LSTM and the decoder has 4 layers of uni-LSTM; the general form of Luong attention~\cite{luong2015effective} is used; both English and Vietnamese word embeddings have 200 dimensions and are randomly initialised. Adam optimiser is used with a learning rate of 0.002 and the maximum gradient norm is set to be 1. If there is a pretraining phase, the learning rate will be halved afterwards. The batch size is 50. Dropout is used with a probability of 0.2. By default, the baseline models are trained with teacher forcing for 60 epochs. The other models are pretrained with teacher forcing for 30 epochs, and then trained with attention forcing for 30 epochs. As discussed previously, the scale $\gamma$ of the attention loss $\mathcal{L}_{\alpha}^{(\tt A)}$ is set to 10, so that its dynamic range is comparable to $\mathcal{L}_{y}^{(\tt A)}$. The default inference approach is beam search; the beam size is 1, in order to reduce the turnaround time. When investigating diversity, sampling search is also adopted, which replaces the selecting operation by sampling. The code for the experiments is available online.\footnote{\url{https://github.com/3dmaisons/af-mnt}}


\subsubsection{Evaluation Metrics}

To measure the overall translation quality and the diversity of the translations, we use the following metrics. 1) BLEU \cite{papineni2002bleu}. The average of 1-to-4 gram BLEU scores are computed and a brevity penalty is applied. The code for computing BLEU score is from the open-source project.\footnote{\url{https://github.com/moses-smt/mosesdecoder}} Higher BLEU indicates higher quality. 2) Pairwise BLEU \cite{shen2019mixture}. For a model $\bm{\theta}$, we use sampling search $M$ times with different random seeds, obtaining a group of translations $\{ \bm{y}'^{(m)} \}_{m=1}^{M}$. $\bm{y}'^{(m)}$ denotes all the output sentences in the dev or test set. Then we compute the average BLEU between all pairs: $\frac{1}{M(M-1)} \sum_{n=1}^{M} \sum_{m=1}^{M} {\tt BLEU} (\bm{y}'^{(n)}, \bm{y}'^{(m)})_{n \neq m}$. In our experiments, $M$ is set to 5. The more diverse the translations, the lower the Pairwise BLEU. 3) Entropy. For a model $\bm{\theta}$, we use beam search, and save the entropy $e_{t}$ of the output token's distribution at each time step. Let $e_{1:T}$ denote all time steps in the dev or test set, we compute the average value: $e = \frac{1}{T} \sum_{t=1}^{T} e_{t}$.  Higher $e$ means that the model is less certain, and thus more likely to produce diverse outputs. This process is deterministic, so there is no need to repeat it with different random seeds.

\subsection{Investigation of Overall Performance}
First, we compare Teacher Forcing (TF) and Vanilla Attention Forcing (VAF). The models are trained respectively with TF and VAF. Our preliminary experiments show that when starting from scratch, the TF model performs much better than the VAF model. When pretraining is adopted, the BLEU of VAF increases from 21.77 to 22.93 for EnFr, and from 13.92 to 18.27 for EnVi. However, it does not outperform TF, as shown by the first two rows in each section of table \ref{tab:BLEU_TF_AAF_tune}. In addition, we observed that after the pretraining phase, when VAF is used, the BLEU decreases. This result is expected, considering the discrete and multi-modal nature of the NMT output space.


\begin{table}
\centering
\begin{tabular}{l|l|l|ll}
\hline
Task  & Training  & $k$     & \multicolumn{1}{l}{BLEU} &              \\
      &  &              & Dev               & Test \\ \hline
EnFr  & TF  & -           & 34.65           & 30.70        \\
      & VAF  & -           & 25.39           & 22.93        \\
      & AAF  & 2.5     & 34.73           & 31.44        \\
      & AAF  & 3.0     & 34.80           & \textbf{31.66}        \\
      & AAF  & 3.5     & 34.71           & 31.34        \\ \hline
EnVi  & TF  & -           & 23.74           & 25.57        \\
      & VAF  & -           & 16.63           & 18.27        \\
      & AAF  & 2.5   & 23.97           & 26.02        \\ 
      & AAF  & 3.0   & 23.81           & 25.71        \\ 
      & AAF  & 3.5   & 23.81           & \textbf{26.72}        \\ \hline
\end{tabular}
\caption{BLEU of TF and AAF with different values of $k$. The higher $k$ is, the more likely the generated output history is used.}
\label{tab:BLEU_TF_AAF_tune}
\end{table}

Next, TF is compared with Automatic Attention Forcing (AAF). To speed up convergence, pretraining is used by default. The hyper parameter $k$, which controls the tendency to use generated outputs, is roughly tuned. The higher $k$ is, the more likely the generated output history is used. Table \ref{tab:BLEU_TF_AAF_tune} shows the BLEU scores of AAF, resulting from different values of $k$. With some simple tuning, AAF outperforms TF. The performance is rather robust in a certain range of $k$. In the following experiments, $k$ is set to 3.0 for EnFr and 3.5 for EnVi.

To reduce the randomness of the experiments, both TF and AAF are run $R=5$ times with different random seeds. Let $\{ \bm{\theta}^{(r)} \}_{r=1}^{R}$ denote the group of TF models, and $\{ \hat{\bm{\theta}}^{(r)} \}_{r=1}^{R}$ the AAF models. For both groups, the BLEU's mean $\pm$ standard deviation is computed. Table \ref{tab:BLEU_TF_AAF} shows the results. For EnFr, AAF yields a consistent 0.44 gain in BLEU. For EnVi, AAF yields a consistent 0.55 gain.

\begin{table}
\centering
\begin{tabular}{l|l|ll}
\hline
Task  & Training     & \multicolumn{1}{l}{BLEU} &              \\
      &              & Dev               & Test \\ \hline
EnFr  & TF           & 34.49$\scriptstyle \pm$0.16           & 31.10$\scriptstyle \pm$0.27        \\
      & AAF     & 34.79$\scriptstyle \pm$0.14           & \textbf{31.54$\scriptstyle \pm$0.14}        \\ \hline
EnVi  & TF           & 23.63$\scriptstyle \pm$0.10           & 25.86$\scriptstyle \pm$0.44        \\
      & AAF   & 23.72$\scriptstyle \pm$0.08              & \textbf{26.41$\scriptstyle \pm$0.33}        \\ \hline
\end{tabular}
\caption{BLEU of TF and AAF, each approach is run 5 times, and the BLEU's mean $\pm$ std is shown. $k=3.0$ for EnFr, and $k=3.5$ for EnVi.}
\label{tab:BLEU_TF_AAF}
\end{table}

\subsection{Investigation of Diversity}
\begin{figure}
\centering
\includegraphics[scale=0.5]{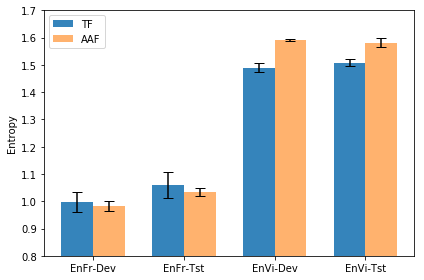}
\caption{Entropy of TF and AAF, each approach is run 5 times, and the entropy's mean $\pm$ std is shown by the height and the error bars. Higher entropy indicates higher diversity.} \label{fig:diverse_entropy}
\end{figure}

\begin{figure}
\centering
\includegraphics[scale=0.5]{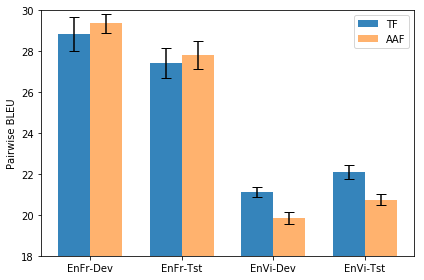}
\caption{Pairwise BLEU of TF and AAF, each approach is run 5 times, and the BLEU's mean $\pm$ std is shown by the height and the error bars. Lower pairwise BLEU indicates higher diversity.} \label{fig:diverse_pbleu}
\end{figure}

To measure the diversity among the translations, the entropy and pairwise BLEU are computed for $\{ \bm{\theta}^{(r)} \}_{r=1}^{R}$ and $\{ \hat{\bm{\theta}}^{(r)} \}_{r=1}^{R}$. Figure \ref{fig:diverse_entropy} shows the entropy resulting from TF and AAF, for both EnFr and EnVi. The height of each bar represents the mean, and the error bars the mean $\pm$ standard deviation. For EnVi, AAF leads to higher entropy, which indicates higher diversity. For EnFr, AAF and TF lead to similar levels of diversity, especially when the standard deviation is considered. We believe that the difference is due to the nature of the tasks. As English and French have similar syntax and lexicon, the models tend to follow the word order of the input sequence, regardless of the training approach. In contrast, English and Vietnamese are more different. When trained with AAF, the model benefits more from using generated back history, which is more diverse than in EnFr. Figure \ref{fig:diverse_pbleu} is the equivalent for pairwise BLEU, and it confirms the above results. Note that pairwise BLEU correlates negatively with diversity. For EnVi, AAF leads to lower pairwise BLEU, i.e. higher diversity. For EnFr, the difference between AAF and TF is negligible. Regardless of the metric, the EnVi translations are more diverse than their EnFr counterpart (higher entropy and lower pairwise BLEU). This also supports the analysis that the difference of the results for EnFr and EnVi is due to the nature of the languages.


\section{Conclusion}
This paper introduces attention forcing for NMT. This approach guides a seq2seq model with generated output history and reference attention, so that the model learns to recover from its mistakes without the need for a schedule or a classifier. Considering the discrete and multi-modal nature of the NMT output space, a selection scheme is added to vanilla attention forcing, which automatically selects a suitable training approach for each pair of training data. Experiments show that attention forcing can improve the overall translation quality and the diversity of the translations, especially when the source and target languages are very different in syntax and lexicon.




\bibliography{af_nmt}

\end{document}